\documentclass[conference]{IEEEtran}

\usepackage[english]{babel}
\usepackage{courier}
\usepackage{amstext}
\usepackage{times}
\usepackage{helvet}
\usepackage{courier}
\usepackage[cmex10]{amsmath}
\usepackage{amssymb}
\usepackage{latexsym}
\usepackage{eurosym}
\usepackage{wasysym}
\usepackage{array}

\makeatletter
\newif\if@restonecol
\makeatother

\usepackage{etex}
\usepackage{tikz}
\usepackage[T1]{fontenc}
\usepackage{pslatex}
\usepackage[arrow, matrix, curve]{xy}
\usepackage{epsfig}
\usepackage{pstricks,egameps}
\usepackage{hyperref, url}

\begin{document}

\title{Research Project:\\ Text Engineering Tool for Ontological Scientometry}

\author{\IEEEauthorblockN{Rustam Tagiew}
\IEEEauthorblockA{Founder of POLAREZ ENGINEERING\\
Alumni of TU Bergakademie Freiberg\\
and of University of Bielefeld\\
yepkio@mail.ru}}

\maketitle

\begin{abstract}
The number of scientific papers grows exponentially in many disciplines. The share of online available papers grows as well. At the same time, the period of time for a paper to loose at chance to be cited anymore shortens. The decay of the citing rate shows similarity to ultradiffusional processes as for other online contents in social networks. The distribution of papers per author shows similarity to the distribution of posts per user in social networks. The rate of uncited papers for online available papers grows while some papers `go viral' in terms of being cited. Summarized, the practice of scientific publishing moves towards the domain of social networks. The goal of this project is to create a text engineering tool, which can semi-automatically categorize a paper according to its type of contribution and extract relationships between them into an ontological database. Semi-automatic categorization means that the mistakes made by automatic pre-categorization and relationship-extraction will be 
corrected 
through a wikipedia-like front-end by volunteers from general public. This tool should not only help researchers and the general public to find relevant supplementary material and peers faster, but also provide more information for research funding agencies. 
\end{abstract}


\IEEEpeerreviewmaketitle

\section{Introduction}
\begin{center}
\begin{minipage}{0.4\textwidth}
 ``\textit{For scientists there are moments, when they read about a paper they were interested in. They are curious about, how the person got exactly to that results and then they find a link from the paper to the data. And then they find out that they can just reproduce the processing of the results immediately. I think that is the step, which a lot of funding agencies pushing towards I know certainly in the US. Pushing to say, if we are funding you, you have to put your data out there, you should put it out there in a standard format, because we paid you to make the data. You may have published a paper, which is based on some results you've noticed. Somebody else may wanna do a very much of a transverse cut for your and anybody else's data. We need to be able to get a reuse of the data we funded.}'' Tim Berners-Lee \cite{semanticweb}
\end{minipage}
\end{center}
\indent There are different types of contributions in science, which all are pressed into text format having title, abstract, body and references. It can be a formula to calculate a sequence of integers for a certain phenomenon. It can be a set of experimentally proven properties for a new chemical element. It can be a machine learning algorithm. It can be an evaluation of a dataset from an experiment. Obviously, some papers originate from a significant practical work beyond reviewing literature and composing the actual text. And papers exist, which do not comprehend such effort. Papers without comprehended practical work either introduce a new theory or are at least inferential from previous papers.\\  
\indent Some of the practical work behind a paper can be directly manifested as digital content, without being presented as a paper. For instance, an algorithm from machine learning can be integrated into machine learning libraries, a chemical formula with a set of properties can be entered into chemical databases, a dataset from a subject research experiment can be uploaded to data repositories and a footage of a ball lightning including its spectrogram \cite{balllightning} can be published on a pod-cast website. This content is also known as {\it supplementary material}. In many cases the supplementary material is of much higher scientific use than the paper text itself. Traditionally, publishing supplementary material alone does not benefit a scientific career and therefore scientists have to go through publishing an obligatory paper.\\  
\indent Most of world research funding and academic position allocation is led by statistics of citations -- present day's scientometry. Published supplementary material still plays a marginal role. This situation is harshly criticized by scientists from diverse disciplines \cite{Lawrence2007,Sengor2014}. The main arguments are the alienation of scientific work from its purpose and the negligence of the practical component. Every scientist has his own representation of publications in his field and his own view on ranking of the relevant research, which does not fit scientometric figures. Since the return to the less transparent, less exact and more time consuming alternative of manual content comparison is not possible, the desire for a better scientometry formed. `Altmetrics' is the present sublimation of this desire and suggests taking more indicators into account than only citations. Those indicators are driven from web data as views and citations in other media.\\
\indent Scientific papers are a content targeted for human readers and therefore noisily formated and require data mining techniques for extraction of statistics. Nevertheless, certain facts are already known \cite{Parolo2015,Evans2008}. The number of scientific papers grows exponentially in many disciplines. The share of online available papers grows as well.  At the same time, the period of time for a paper to loose at chance to be cited anymore shortens. The decay of citing rate shows similarity to ultradiffusional processes as for other online contents in social networks. The distribution of papers per author shows similarity to the distribution of posts per user in social networks. The rate of uncited papers for online available papers grows while some papers `go viral' in terms of being cited. Summarized, the practice of scientific research moves towards the domain of unstructured social networks vulnerable to hypes and content repetition.\\ 
\indent There is a lot of effort in all disciplines to create clear taxonomy of research subjects and to introduce standard keywords. Nevertheless, computer scientists from the `neuronal nets' community reinvent approaches from the `optimization' community and economists redo experiments from psychology. Given the exponential growth at number of papers, researchers in different categories may end up doing the same thing without knowing about each other. Researchers use different words to describe the same semantics -- the practical work they conducted. The sheer exponentially growing number of publications requires more and more of automated assistance in semantical analysis of the publications, in their structuring and in guidance of researchers and funding agencies.\\
\section{Terminology}
\indent A piece of scientific research, which is published, is a \textit{publication}. It can be a \textit{paper}, which is of format having title, abstract, body and references. It can also be a dataset, a coded algorithm or an audiovisual sample -- \textit{suplementary material}. It can also be a combination of these things. A \text{scientific document} is a paper in any device independent file format like PDF, PS, DVI and so on. The \textit{metadata} of paper is represented by its BIBTEX entry plus the list of cited publications.\\   
\section{Description of the Objective}
\begin{figure}[t]
\begin{center}
\includegraphics[scale=0.45]{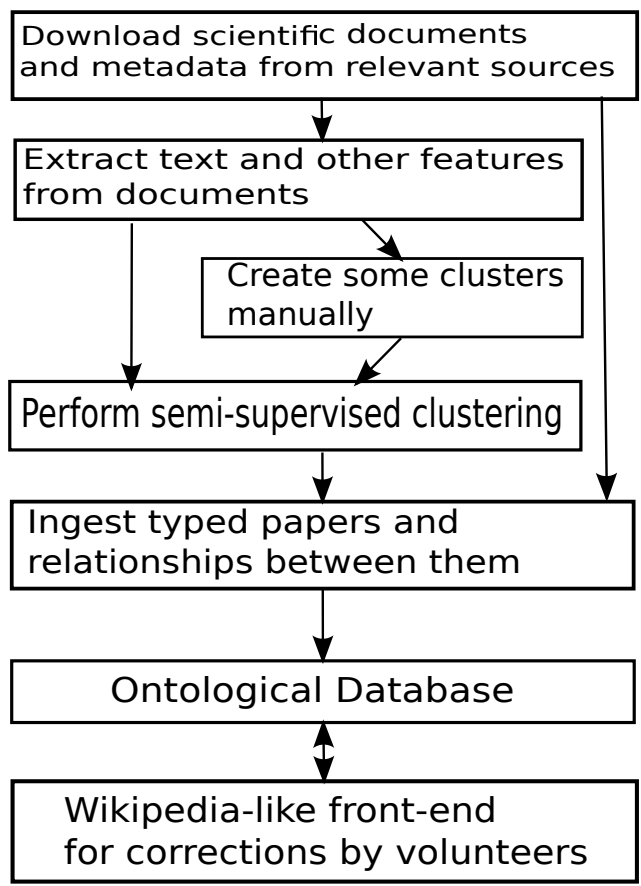}
\caption{The design of the system. The document and meta data sources are Citeseerx, Mendeley and so on. Text and feature extraction and the method of clustering will be evaluated on the test set of manually made cluster. The wikipedia-like front-end can be a customizing Drupal system.}
\label{successmm}
\end{center}
\end{figure}
\indent The objective is to represent every scientist's knowledge about practical research in his field as an ontology assisted by text classification. There are two sides of semantics for this ontology -- the human and the machine one. The human semantics is the description of research and the practical work behind it. The machine semantics is a document classifier and a set of rules for processing of metadata into relationships between papers.\\
\indent The design of the semi-automatically created ontological database for scientific documents is relatively simple, as you can see in Fig.\ref{successmm}. Scientific documents and their metadata are prepared and offered under conditions by many commercial and non-commercial entities such as Citeseerx, Mendeley, Thomas Reuters (TR) web of science, Scopus and ResearchGate. There are also domain-specific repositories such as DBLP and RePEc/IDEAS. The preparation of metadata is by no means trivial \cite{Wu2014}, leads to noisy results, and requires its own scientific community in between of `information retrieval' and `scientometrics' \cite{Mayr2014}.\\
\indent From the acquired documents, text and features are to be extracted and the documents are to be clustered on their base. The method of clustering is to be semi-supervised \cite{Aggarwal2012} -- the clustering results are to be evaluated on a partial set of manually made sample clusters. Unsupervised clustering may lead to unwanted results such as clusters of different proficiency in english e.g.. Creating a full set of sample clusters for every type of paper would require too much of manual work For the supervised classification as we will see in section \ref{manually}. The results of cluster evaluation will reflect on text/feature extraction as well as on clustering method.\\
\indent Recent development of open source text engineering tools is promising and will play a crucial role in the task of clustering. For instance, GATE (general architecture for text engineering) is a free open source software agglomerate \cite{gate}, which includes libraries such as KEA (keyphrase extraction) \cite{kea}, which in turn uses WEKA (Waikato data mining library) \cite{wekalib}, which is an aggregation of machine learning algorithms.\\
\indent Once the types of the papers are determined, the metadata can tell much more than just (co)citation and (co)authorship. Now the type of a citation can defined. For instance, when a computer science paper evaluating algorithms cites a paper about chemical reactions, the type of citation will be different to the case, where another chemical paper cites it. In the first case the paper is cited for its dataset, and in the second, for its insight. The application of formal concept analysis \cite{PoelmansIKD14} will be considered as a framework for the relationships between the types of papers.\\
\indent One can expect that the information ingested into the ontological database will contain mistakes. For instance, some papers will be assigned to wrong types. This problem is easily solved by wikipedia-like front-end. Like on Wikipedia, volunteers will correct the database. A subset of volunteers would be the researchers themselves, because some of them are interested in having their research being appropriately presented. The system to be chosen for such a front-end is yet to be chosen from these available open source.\\  
\indent Although the design is simple, the amount of its purposes is huge. The ontological database should be able to answer questions on the relationships of a certain papers to other papers. It should also be able to answer the question about the type of supplementary material the authors could be asked for. It should allow machines to communicate with and guide scientists in a more detailed way than just displaying them their h-index \cite{Hirsch:2005zc}.\\
\indent The main scientific contribution lies in the development of the scientometric ontology and its human and machine semantics. The development of machine semantics implies contributions in the domain of document clustering.\\ 
\section{Accessing Supplementary Material}
\indent In scientific tradition supplementary material of a paper can be accessed by sending a request to the authors. This tradition is based on the idea that any insight should be reviewed by colleagues based on supplementary material. Having a scientometric ontology, where every paper has a type, one would easily form a query to get names of authors, which have a similar supplementary material, and address them all in the request for the supplementary materials.\\ 
\indent The less traditional way is the publication of the supplementary material together with the paper. The second approach obviously provides a faster access. The authors hereby have to agree with a certain type of license, whose disadvantages may overweight the academic incentive for some of the authors.\\
\indent For standalone publications of datasets as a subtype of supplementary material, the academic incentive is recently declared by Data Citation Synthesis Group through a list of justifying reasons \cite{datacitation}. VegBank \cite{vegbank} is an example for an online repository, where obligatory citable datasets of a certain type are stored. A scientometric ontology from the realm of papers accompanied by supplementary materials would provide a reasonable structure for standalone material publications.\\  
\section{Creating Clusters Manually}
\label{manually}
\begin{figure}[t]
\begin{center}
\includegraphics[scale=0.45]{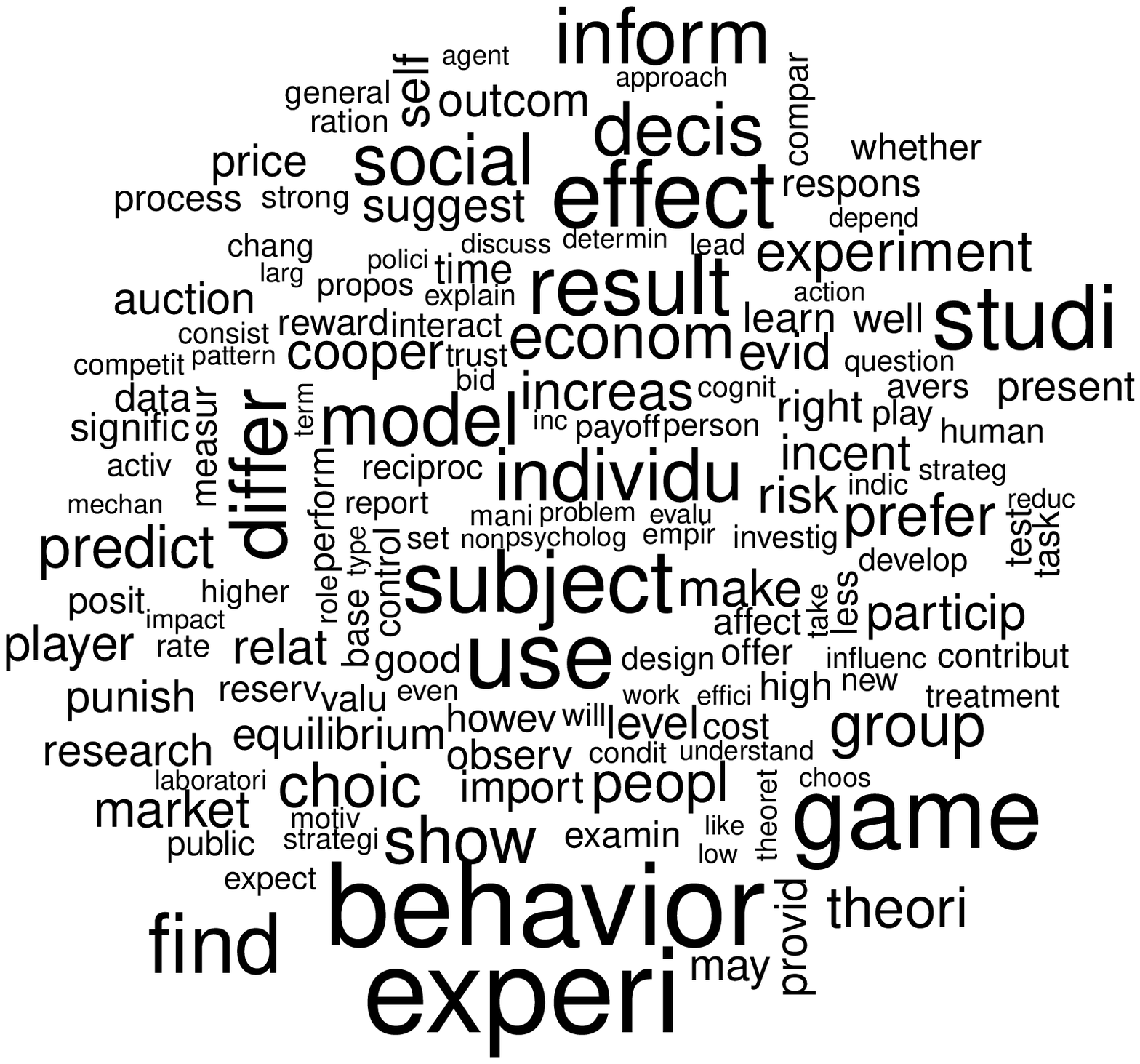}
\includegraphics[scale=0.45]{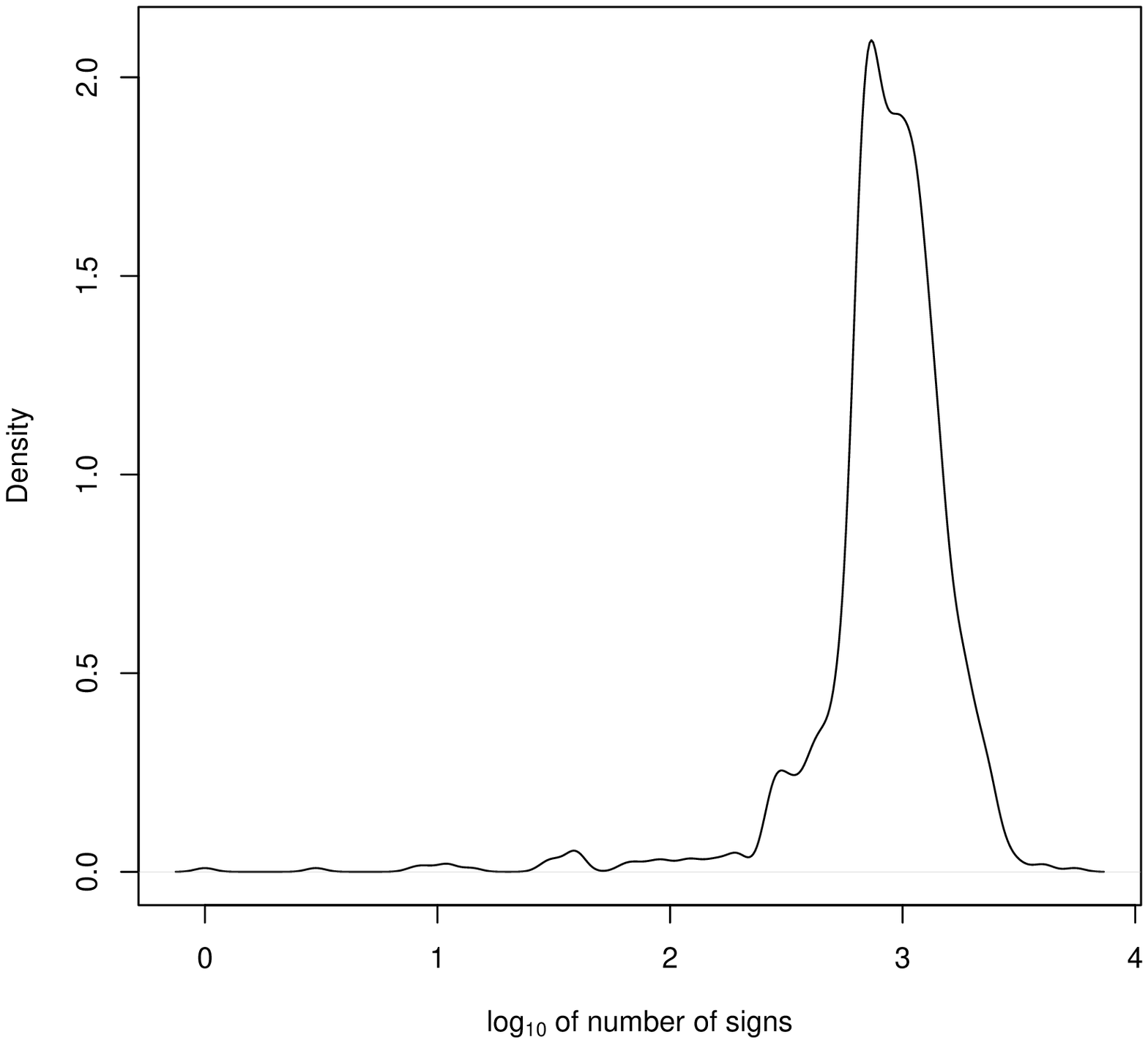}
\caption{Top: The word stem cloud of abstracts for a sample cluster of documents, which have datasets from laboratory behavior studies as supplementary material. Bottom: The distribution of number of signs in an abstract as parsed by Mendeley.}
\label{wordcloud}
\end{center}
\end{figure}

\indent Creating clusters manually is a tricky and very domain dependent procedure. We present it on two examples.\\  
\indent The first example is about human behavior and decision making research. There are two different communities, which struggle with two sides of the same problem. Those are the data scientists dealing with huge noisy datasets of human behavior from web and behavioral economists gathering clean small but insightful datasets while conducting experiments on human subjects \cite{tia}. The datasets of data scientists are mostly confidential and results are not always published online. There is a certain interest to integrate the domain knowledge and datasets from behavioral economics into web mining. Since the website ExLab (exlab.bus.ucf.edu) for supplementary materials of experimental human subject research is defunct, there is no other way to accessing the data than to email the authors. One can define a new type `laboratory behavior study'$=$\textit{Labbehavior} in the scientometric ontology. Labbehavior has following description in natural language:
\begin{center}
\begin{minipage}{0.4\textwidth}
\textit{It is a publication with experimental behavioral data being involved. It it should not be a paper based only on poll data -- subject actions should be recorded and not assumed based on poll ratings. The subjects are humans or higher primates -- no studies with computer agents participating all along. The indicated discipline is either economics, psychology, social sciences or computer science. It is neither a pure summary paper nor a paper based on a pure field study nor on a pure field experiment. It should not be a study of cognitive abilities. It can be a study of behavior under bio-chemical influence such as through hormones and drugs. It should not be a pure game theoretical paper.}
\end{minipage}
\end{center}
\indent Now, let us create a set of objects of type Labbehavior. First step would be to get the names of main authors of behavioral economics from Wikipedia. Here are the names of the authors:  
\begin{center}
\begin{minipage}{0.4\textwidth}
\textit{Ernst Fehr, Dan Ariely, Urs Fischbacher, Colin Camerer, Simon G\"achter, Amos Tversky, Armin Falk, Reinhard Selten, Daniel Kahneman, Uri Gneezy, B. Douglas Bernheim, George Loewenstein, Matthew Rabin,  Herbert A. Simon, Vernon L. Smith, Larry Summers, Michael Taillard, Richard Thaler, John Quiggin, Margaret McConne, Werner De Bondt, Roy Baumeister, Ed Diener, Ward Edwards, Gerd Gigerenzer, George Katona, Steven Lea, Walter Mischel, Drazen Prelec, Paul Slovic, Malcolm Baker, Nicholas Barberis, Gunduz Caginalp, David Hirshleifer, Andrew Lo, Michael Mauboussin, Terrance Odean, Richard L. Peterson, Charles Plott, Hersh Shefrin, Robert Shille, Andrei Shleifer, Robert Vishny}
\end{minipage}
\end{center}
\indent Then we add about 500 names from the z-Tree emailing list. z-Tree is a popular software for conducting experiments. Having this list of names, Google Scholar and Citeseerx can be used to crawl the documents, mostly in the PDF format. The crawled documents are the top 10-40 results for every name and first 100 for the term `z-Tree'. Running Mendeley on these documents gives a list of \textit{noisy} BIBTEX entries including the abstracts. `Noisy' means random mistakes in the entries. The results of the combination pdftotext \cite{pdftotext} and the ParsCit library \cite{parscit} are noisier. Abstracts are not always sufficient to classify a paper -- a closer look into the documents is sometimes required. Any volunteer for manual sorting should be warned -- this task requires a firm control of own attention, since every paper can be a surprise and distract attention to further reading. It is like sorting a library. For instance, papers on negotiation skills under influence of drugs and even papers on 
supernatural powers appeared among the crawled documents. 
Fig.\ref{wordcloud} shows a word stem cloud of the abstracts in the created Labbehaviour cluster of ca. 1000 documents. The results of application of pdftotext on documents from this cluster can be downloaded from here: \href{https://copy.com/WQc7UO9ICzorxAzz}{copy.com/WQc7UO9ICzorxAzz}.\\
\indent The second example is from computer science. For the discipline of machine learning, the papers describing algorithms form the type \textit{MLalgo}. BIBTEX entries of ca. 100 such papers can be easily derived from the source code of WEKA library. The results of application of pdftotext on documents from this cluster can be downloaded from here:  \href{https://copy.com/OBnaf11bWnL4LcQm}{copy.com/OBnaf11bWnL4LcQm}.\\  
\section{Related Work}
\indent Let us call the type for the related papers to the clustering part of this research proposal as \textit{ScienceDocCl}. The semantics of ScienceDocCl is:
\begin{center}
\begin{minipage}{0.4\textwidth}
\textit{It is a publication with clusters being calculated on scientific documents. The presented method should be scalable up to 5M of papers as currently available on Citeseerx. It should not be clustering based on (co)citations or other metadata. The clusters of scientific documents should be formed according to the type of performed practical work, and not according to the topic of research.}
\end{minipage}
\end{center}
\indent Evaluating clustering algorithms on sets of documents has been a research subject since decades \cite{Aggarwal2012}. One paper \cite{Boyack2011} is especially interesting, since it evaluates clustering algorithms on 2M of scientific documents. The best result of clustering could be impressively depicted by graph visualization software. Unfortunately, it was a topic oriented clustering. Clustering according to the type of performed practical work will be a bigger challenge -- it requires a deeper understanding of the language, than bag of words statistics.\\
\indent The website of semantic web journal \cite{sematicwebjournal, Hu2013} presents a related work to the wikipedia-like web-interface upon an ontological database. It is indeed a scientometric ontology with information about the peer-review process for every paper. Nevertheless, there is no way to correct or augment this knowledge base by internet users.\\
\section{Time \& Cost Plan}
\indent The main practical work for this research proposal will be represented by building a `Big Data' infrastructure, massively (24/7) lunching parallel runs with diverse clustering approaches and recording the performance of generated classifiers on the manually created sample clusters. This would require a certain investition into hardware. There is a certain trend towards GPU in text clustering \cite{Zhang2011}.\\
\indent The collection of scientific documents and their metadata is not a fully trivial task. There is a cheap way of downloading that data from Citeseerx, which will give very noisy metadata. And there is a more expensive way of buying cleaner data from sources like Thomas Reuters (TR) web of science.\\
\indent Finally, a quest for a user-edited ontology website system will run in parallel. Regarding the current pace of progress, such systems will be developed in a year or two.\\ 
\bibliographystyle{IEEEtran}
\bibliography{webmine01}
\end{document}